\documentclass[letterpaper, 10 pt, conference]{ieeeconf}  

\IEEEoverridecommandlockouts                              

\usepackage{amsmath}
\usepackage{amssymb}
\usepackage{graphicx}
\usepackage[Symbol]{upgreek}
\usepackage{epstopdf}
\usepackage{subcaption}
\usepackage{enumerate}
\usepackage{paralist}

\usepackage{float}

\usepackage{latexsym}
\usepackage{multicol}
\usepackage{multirow}
\usepackage{lipsum}

\usepackage{cite}

\usepackage{makecell}
\usepackage{breqn}

\usepackage{verbatim}

\usepackage{color}
\usepackage{xcolor}
\usepackage{bm}
\usepackage[font=small,labelfont=bf]{caption}

\usepackage{courier}
\usepackage{tikz}
\usetikzlibrary{calc}

\newtheorem{remark}{Remark}

\newcommand{\vect}[1]{\bm{#1}}

\newcommand{\bvect}[1]{\bar{\vect{#1}}}

\newcommand{\dvect}[1]{\dot{\vect{#1}}}
\newcommand{\ddvect}[1]{\ddot{\vect{#1}}}

\newcommand{\vphi}[0]{\vect{\phi}}

\newcommand{\partDer}[2]{\frac{\partial #1}{\partial #2}}
\newcommand{\partDerSq}[2]{\frac{\partial^2 #1}{\partial^2 #2}}

\def\particulartemplate#1{
  \begin{tikzpicture}[overlay, remember picture]
    \draw let \p1 = (current page.west), \p2 = (current page.east) in
      node[minimum width=\x2-\x1, minimum height=0.1cm, rectangle, fill=yellow!35!white, anchor=north west, align=center, text width=\x2-\x1] at ($(current page.north west) + (0,-0.3)$) {\large \textbf{\texttt{#1}} };
  \end{tikzpicture}
}

\title{\LARGE \bf
A Reversible Dynamic Movement Primitive formulation
}

\author{Antonis Sidiropoulos$^{1}$ and Zoe Doulgeri$^{2}$
\thanks{The research leading to these results has received funding by the EU Horizon 2020 Research and Innovation Programme under grant agreement No 820767, project CoLLaboratE.}
\thanks{$^{1,2}$Aristotle University of Thessaloniki, Department of
Electrical and Computer Engineering, Thessaloniki 54124, Greece. 
        {\tt\small antosidi@ece.auth.gr, doulgeri@eng.auth.gr}}%
}

\begin{document}

\maketitle

\particulartemplate{
This paper is a Post-Print version (i.e. final draft post-refereeing). \\
The publisher's version can be accessed through \\
DOI: 10.1109/ICRA48506.2021.9562059
}


\thispagestyle{empty}
\pagestyle{empty}

\begin{abstract}
    In this work, a novel Dynamic Movement Primitive (DMP) formulation is proposed which supports reversibility, i.e. backwards reproduction of a learned trajectory.
    Apart from sharing all favourable properties of the original DMP, decoupling the teaching of position and velocity profiles and bidirectional drivability along the encoded path are also supported. 
    Original DMP have been extensively used for encoding and reproducing a desired motion pattern in several robotic applications. However, they lack reversibility, which is a useful and expedient property that can be leveraged in many scenarios. The proposed formulation is analyzed theoretically and its practical usefulness is showcased in an assembly by insertion experimental scenario.
\end{abstract}

\section{Introduction} \label{sec:Introduction}

The technological advancements in the field of robotics over the past few years have triggered a surge in research interest for incorporating robots more and more actively in everyday life. 
However, programming a robot to perform even simple tasks usually requires considerable amount of time and advanced robotic and programming knowledge. To facilitate this cause, the use of Programming by Demonstration (PbD) has been proposed for passing on human skills to a robot \cite{PbD_Billard}. At the core of this approach is a model used for encoding the demonstrated trajectory encapsulating the desired motion pattern for performing a task. The properties of this model play a major role, as they determine whether the encoded motion is amenable to generalization, online adaption, reverse execution etc, depending on the target application.

In the literature there are a lot of approaches for encoding a desired motion pattern, such as spline decomposition \cite{Splines_2006}, neural networks \cite{NN_2018}, Hidden Markov Models \cite{HMM_Calinon_2010}, Gaussian Process Regression \cite{GPR_2008}, Gaussian Mixture Models \cite{GMM_Zadeh_2010} and more. Undoubtedly, one of the most prominent approaches is the Dynamic Movement Primitives which have been applied in numerous robotic applications \cite{Pastor_2011_DMP, DMP_Pastor_2009, DMP_Mulling_2013, DMP_Hirche_2014}. The reason behind this widespread applicability of DMP is mainly attributed to the favourable properties they come along with. Specifically, they offer a fairly simple framework for encoding any smooth non-linear trajectory, allowing generalization to new targets (end points) and changing the speed of execution. Moreover, they guarantee global asymptotic stability (GAS) at the target, robustness to target perturbations and allow seamlessly the encoding and synchronization of multiple DoFs. All these properties make DMP quite the appealing and efficient choice to opt for when encoding a desired motion pattern. However, there is still room and need for improvement as original DMP do not support reversibility. There are many tasks where reversibility, i.e. reproduction of the learned trajectory backwards in time, could come quite in handy.

For many tasks, reverse execution could enable automatic derivation of certain required operations from their forwards counterparts. For instance, in assembly tasks a reversible model could be used for assembling and disassembling a part. In general, when performing any task that involves reaching a target, like a handover, a placing of a part etc., the reverse motion could be used for the retraction, saving the effort of explicitly encoding and programming the backward motion. This becomes even more useful in tasks executed in cluttered environments. Reversibility can also prove really effective and practical for recovering from errors during the execution of a task. These errors could originate from faulty or noisy sensor measurements or could be attributed to disturbances. Using reverse execution, the robot can temporarily back out of an erroneous situation by tracing back to a previous point, after which the execution can be automatically retried. There are some works that have applied reversibility to recover from such errors in practical applications \cite{Towards_Reversible_DSL, Auto_error_recovery_reverse_exec}. In these works, the authors address error recovery by reversing the program's execution by means of designing a domain specific language (DSL) which supports running backwards the executed programming commands.  Reversibility using DMP is addressed in \cite{Towards_Rev_DMP} with a DMP formulation which however fails to ensure global asymptotic stability (GAS) and reversibility. Eventually the authors resort to the use of two separate DMPs to ensure both. Two separate DMPs to achieve reversibility are also used in \cite{Nemec_Rev_DMP_2019}.

In this work, we propose a novel DMP formulation that achieves reversibility as opposed to the original DMP, maintaining  all its favourable properties. Moreover, it allows a two phase PbD 
for velocity profile teaching. Learning the velocity profile in a second phase is particularly suitable for encoding tasks that require high precision during the demonstration of the desired path. The proposed formulation allows not only replaying the trajectory back from its goal, but also moving back and forth along the learned path anywhere during the execution.
Other qualities of the proposed formulation is that training requires only position measurements and is decoupled from the DMP's stiffness and damping gains. Additionally, the DMP's effective stiffness and damping is not dependent on the temporal scaling parameter.    
The proposed DMP is theoretically analyzed and compared with the original one.
Its benefits are demonstrated experimentally in an assembly by  insertion scenario inspired by a real case.

The rest of this paper is organized as follows: section \ref{sec:DMP_prelim} provides an overview on original DMP. In section \ref{sec:novel_DMP} the proposed DMP formulation is presented and its properties are detailed in section \ref{sec:Comparison}. Experimental results are presented in section \ref{sec:Experiments} and  conclusions are drawn in section \ref{sec:Conclusions}. In the appendix, the correspondence between the proposed and original DMP formulations is analyzed and preliminaries on unit quaternions are provided.


\section{Dynamic Movement Primitives Preliminaries} \label{sec:DMP_prelim}

DMP consist of a transformation system which generates a trajectory and a canonical system for controlling the system's temporal evolution \cite{Ijspeert2013}. The transformation system is composed of a second order linear attractor to a goal and a non-linear forcing term  which has to be learned in order to encode a desired kinematic behavior. 
Aggregating the different variants that have been proposed in the literature in a more generic formulation
a $1$-DoF DMP is given as follows:
\begin{align} 
    &\tau^2 \ddot{y} = \alpha_z \beta_z (g - y) - \alpha_z \tau \dot{y} + g_f(x)(g-y_0)f_s(x) \label{eq:DMP_y_ddot} \\
    &\tau \dot{x} = h(x) \label{eq:DMP_x_dot}
\end{align}
or expressing  \eqref{eq:DMP_y_ddot} in state equations introducing $z = \tau \dot{y}$:
\begin{align} 
    &\tau \dot{z} = \alpha_z \beta_z (g - y) - \alpha_z z + g_f(x)(g-y_0)f_s(x) \label{eq:DMP_z_dot} \\
    &\tau \dot{y} = z \label{eq:DMP_y_dot}
\end{align}
where $y, \dot{y}$ is the position and velocity, $g$ the target, $y_0$ the initial position. The phase variable $x$ is used to avoid direct dependency on time and $\tau>0$ is a temporal scaling parameter, typically set equal to the movement's total duration $T = t_f - t_0$, where $t_0$ is the initial and $t_f$ the final time instant.
The forcing term $f_s(x)$ is given by:
\begin{equation} \label{eq:DMP_f_s}
    f_s(x) = \vphi(x)^T \vect{w}
\end{equation}
where $\vphi(x)^T = [\psi_1(x) \ \cdots \ \psi_N(x)]/{\sum_{i=1}^{N} \psi_i(x)}$, with $\psi_i(x) = exp(-h_i(x-c_i)^2)$. The gating function $g_f(x)$ 
ensures that the forcing term eventually vanishes thus \eqref{eq:DMP_y_ddot} acts as a spring-damper and converges asymptotically to the goal $g$.

The canonical system's evolution is determined by  \eqref{eq:DMP_x_dot}, with $h(x)$ chosen so that $x$ evolves monotonically from its initial value $x(0) = x_0$ to its final value $x(t_f) = x_f$. Many options are available, e.g. a linear canonical system $\tau \dot{x} = 1$ for $x\le1$ and $\tau \dot{x}=0$ otherwise, with $x_0 = 0$ and $x_f=1$; an alternative choice is an exponential one $\tau \dot{x} = -a_x x$ with $x_0 = 1$ and $x_f=0^+$. For the gating function $g_f(x)$ there are many choices as well, such as linear, exponential or sigmoid gating, to name but a few.
Regarding the choice of $\alpha_z, \beta_z$ a typical approach is to set $\beta_z = \alpha_z/4>0$ to render the linear part of \eqref{eq:DMP_y_ddot} critically damped.
To achieve good approximation during learning, a general heuristic is to place the centers $c_i$ of the Gaussian kernels in \eqref{eq:DMP_f_s} equally spaced in time. 
A typical choice is to then set the inverse widths of the Gaussians as $h_i = \frac{a_h}{(c_{i+1}-c_i)^2}$, $h_N = h_{N-1}$, $i=1,\cdots,N$, where $a_h > 0$ is a scaling factor controlling the overlapping between the kernels. 

\section{A Reversible Dynamic Movement Primitive} \label{sec:novel_DMP}
Reversibility of DMP means that the generated system's trajectory will be identical in forward and backward motion.
Given a goal attractor dynamics e.g. (1), it is however impossible to have global stability and reversibility simultaneously as also noted in \cite{Towards_Rev_DMP}. Original DMP are indeed not reversible as shown at the end of Appendix A. In order to produce a reversible DMP the authors in \cite{Towards_Rev_DMP} suggest a non-linear system with two equilibria, an attractor and a repeller, and build their proposed formulation around it. This formulation achieves  partial reversibility if the solution stays within a region and becomes unstable beyond. As the solution cannot be guaranteed to stay within the permitted region, which is expected as the basic dynamics are non-linear, the authors finally resort to using two forcing terms to ensure global stability. 
Instead of having a goal attractor as a foundation of the transformation system we base our proposed formulation on a linear system with a global asymptotically stable origin. Inspired by linear trajectory tracking dynamics we propose the following DMP structure for $1$-DoF:
\begin{align}
    \ddot{y} &= \ddot{y}_{x} - D(\dot{y}-\dot{y}_{x}) - K(y - y_{x} ) \label{eq:GMP_imp_form} \\
    \dot{x} &= h(x)/\tau \ , \ x(t_0 / \tau) = x_0 \ , \ x(t_f / \tau) = x_f \label{eq:GMP_x_dot}
\end{align}
where $K,D$ are positive scalars, $h(x)$ evolves monotonically from $x_0$ to $x_f$ and must be zero for $x$ outside of the interval defined by $x_0$ and $x_f$ 
(the canonical system examples presented in Section \ref{sec:DMP_prelim} are appropriate choices); lastly, $y_x$ is the desired spatially and temporally scaled trajectory, generated by:
\begin{align} 
    &y_{x} \triangleq k_s(f_p(x) - f_p(x_0)) + y_0 \label{eq:GMP_y_x} \\
    &\dot{y}_{x}(t) = k_s\dot{f}_p(x) \label{eq:GMP_y_x_dot} \\
    &\ddot{y}_{x}(t) = k_s\ddot{f}_p(x) \label{eq:GMP_y_x_ddot}
\end{align}
where 
\begin{equation} \label{eq:GMP_ks}
    k_s \triangleq \frac{g-y_0}{f_p(x_f) - f_p(x_0)}
\end{equation}
is the spatial scaling term and
\begin{equation} \label{eq:GMP_fp}
    f_p(x) = \vphi(x)^T \vect{w}
\end{equation}
encodes the demonstrated position trajectory through a weighted sum of Gaussians,
with $w = \text{argmin}_w J(f_p(x), y_d)$, where $J$ can be any objective cost function, e.g. the Least Squares (LS) or the Locally Weighted Regression (LWR) objective \cite{Ijspeert2013}.
The derivatives of $f_p(x)$ can be calculated analytically from $\dot{f}_p(x) = (\frac{\partial \vphi}{\partial x}\dot{x})^T\vect{w}$ and $\ddot{f}_p(x) = (\frac{\partial^2 \vphi}{\partial x^2} \dot{x}^2 + \frac{\partial \vphi}{\partial x} \ddot{x})^T\vect{w}$.

Forward reproduction of the learned trajectory is accomplished integrating \eqref{eq:GMP_imp_form}-\eqref{eq:GMP_x_dot}. For backward reproduction only the sign of $\dot{x}$ needs to be flipped, i.e. $\dot{x} = -h(x) / \tau$, $x(t_0/\tau)=x_f$.

Denoting $e = y - y_x$ and considering that $D,K>0$ it follows from \eqref{eq:GMP_imp_form} that $e,\dot{e},\ddot{e} \rightarrow 0$. What is more, as  $y_x$ and its derivatives are bounded by construction, $y,\dot{y},\ddot{y}$ are bounded as well. In forward reproduction, (7) implies that $x \rightarrow x_f$ and $\dot{x},\ddot{x} \rightarrow 0$, and from \eqref{eq:GMP_y_x}-\eqref{eq:GMP_y_x_ddot} and \eqref{eq:GMP_ks} we can conclude that $y_x \rightarrow g$, $\dot{y}_x,\ddot{y}_x \rightarrow 0$. Therefore, $y \rightarrow  g$, $\dot{y},\ddot{y} \rightarrow  0$.
In backward reproduction from any $x \ne x_0$,  $\dot{x} = -h(x) / \tau$ implies $x \rightarrow x_0$ and $\dot{x},\ddot{x} \rightarrow 0$, which from \eqref{eq:GMP_y_x}-\eqref{eq:GMP_y_x_ddot} and \eqref{eq:GMP_ks} implies that $y_x \rightarrow y_0$, $\dot{y}_x,\ddot{y}_x \rightarrow 0$, hence $y \rightarrow  y_0$, $\dot{y},\ddot{y} \rightarrow  0$.

Notice that \eqref{eq:GMP_imp_form} can also be written as $\ddot{y} = K(g - y) - D\dot{y} + f'(x)$, where $f'(x) = \ddot{y}_{x} + D\dot{y}_{x} - K(g - y_{x})$, which is a linear system modulated by the nonlinear  forcing term $f'(x)$ which fades to zero as $t \rightarrow \infty$ and consequently it has a GAS point attractor  like the original DMP.

For Cartesian position encoding the above formulation is readily extended as each DoF can be encoded separately and synchronization can be achieved using the same canonical system \eqref{eq:GMP_x_dot} for all DoFs.
Regarding Cartesian orientation, a formulation analogous to \eqref{eq:GMP_imp_form} is adopted, which makes use of the quaternion logarithm as in \cite{DMP_orient_Koutras} (a brief summary on unit quaternions is provided in Appendix B). Introducing $\vect{\eta} \triangleq \log(\vect{Q}*\bvect{Q}_0)$, where $\vect{Q}$ is the current, $\vect{Q}_0$ the initial orientation expressed  as unit quaternions, $*$ denotes the quaternion product, $\bar{(.)}$ the quaternion inverse and $\log(.)$ the quaternion logarithm, the proposed DMP formulation for Cartesian orientation is given by:
\begin{equation} \label{eq:GMP_orient_ddq}
     \ddvect{\eta} = \ddvect{\eta}_x - \vect{D}(\dvect{\eta} - \dvect{\eta}_x) - \vect{K}(\vect{\eta} - \vect{\eta}_x)
\end{equation}
with the canonical system given by \eqref{eq:GMP_x_dot}.
The matrices $\vect{K}, \ \vect{D}$ are diagonal positive definite stiffness and damping matrices. The desired spatially and temporally scaled trajectory is given by:
\begin{equation} \label{eq:GMP_orient_q_ref}
    \vect{\eta}_{x} \triangleq \vect{K}_{s} \vect{f}_q(x)
\end{equation}
where $\vect{K}_s \triangleq diag\left( \vect{\eta}_g ./ \vect{f}_q(x_f) \right)$ with $\vect{\eta}_g \triangleq \log(\vect{Q}_g*\bvect{Q}_0)$, $\vect{Q}_g$ is the target orientation and $./$ denotes the element-wise division.
The desired motion is learned based only on the demonstrated orientation $\vect{\eta}_d = \log(\vect{Q}_d*\bvect{Q}_{d,0})$, where $\vect{Q}_d$ and $\vect{Q}_{d,0}$ are the demonstrated  and initial orientation, through $\vect{f}_q(x)$ which is a weighted sum of Gaussians:
\begin{equation}
    \vect{f}_q(x) = \vect{W}\vphi(x)
\end{equation}
where $\vect{W} = [\vect{w}_x \ \vect{w}_y \ \vect{w}_z]^T$, with $\vect{w}_i$ being the weights for each coordinate $i \in \{x,y,z\}$.
The desired scaled velocity $\dvect{\eta}_{x}$ and acceleration $\ddot{\vect{\eta}}_{x}$ are obtained by differentiating $\vect{\eta}_{x}(t)$ as in \eqref{eq:GMP_y_x_dot}, \eqref{eq:GMP_y_x_ddot}.
The orientation  $\vect{Q}$, rotational velocity, $\vect{\omega}$ and acceleration $\dvect{\omega}$ can be obtained from $\vect{\eta}$, $\dvect{\eta}$, $\ddvect{\eta}$ using $\vect{Q} = \exp(\vect{\eta})*\vect{Q}_0$ and \eqref{eq:dq_omega}, \eqref{eq:dvRot_ddq} from Appendix B.
Finally, the stabilty analysis for 1-DoF is readily extended to the orientation formulation \eqref{eq:GMP_orient_ddq}, as each DoF in $\vect{\eta}$ is decoupled.


\section{Proposed DMP Properties} \label{sec:Comparison}

\subsection{Standard DMP properties}

The proposed DMP formulation retains all desirable properties of the original DMP. 
Global asymptotic stability at the target has already been discussed in the previous section. 
It is also clear that the stable coordination of multiple DoFs is similar to the original DMP, using a common canonical system for all DoFs. 

Spatial and temporal scaling of the novel DMP, i.e. generation of trajectories that are qualitatively similar or topologically equivalent \cite{Ijspeert2013} when there is a change in the initial/target position or the temporal scaling parameter $\tau$, follows straightforwardly from the mathematical correspondence between the two as detailed in Appendix A. It can also be easily verified through simulations that for different initial/target poses or temporal scaling, the trajectories produced by the proposed DMP coincide with those of the original one.

Robustness to perturbations is also a trait of the proposed formulation. Such perturbations could be for instance an external disturbance. In this case, as in the original DMP, the phase stopping mechanism is employed, which results in the slow down or even halt of the trajectory generation by modifying the canonical system's evolution. Denoting the disturbance by $d(t)$, phase stopping can be implemented by modifying the canonical system's evolution as follows:
\begin{equation} \label{eq:phase_stop}
    \tau \dot{x} = \frac{h(x)}{1 + a_d|d(t)|}
\end{equation}
with $a_d>0$. Other phase stopping types are also possible, like a sigmoid stopping \cite{Vlachos}.
Another type of perturbation is the change of the target position on the fly. The DMP will modulate its trajectory to reach the new target.

Another appealing property of DMP is the capability of incorporating coupling terms, so as to modify online the dynamical system's trajectory based on external signals, obviating the need of trajectory replanning. Coupling terms have been applied to adjust online the DMP's trajectory based on the external force measurements \cite{Pastor_2011_DMP}, to enforce position/joint limits \cite{DMP_Gams_2009}, for obstacle avoidance \cite{Bio_DMP_2009} and other. These coupling terms, can also be included in the proposed DMP, to attain the desired behaviour.
For limit avoidance, we can introduce the state $z = \dot{y}$, to rewrite the novel DMP in the form of state equations and then add the repulsive force at the velocity level as in the original DMP:
\begin{align*}
    \dvect{y} &= \vect{z} - \frac{\gamma}{(\vect{y}_L - \vect{y})^3} \\
    \dvect{z} &= \ddvect{y}_x -\vect{D}(\vect{z} - \dvect{y}_x) - \vect{K}(\vect{y}-\vect{y}_x)
\end{align*}
where $\vect{y}_L$ is the limit and $\gamma>0$ controls the effect of the repulsive force.
Obstacle avoidance can also be achieved as in \cite{Bio_DMP_2009}.
Denoting $\vect{p}_o$ the obstacle's position, the term $ \vect{f}_o = \gamma \vect{R} \dvect{y} \phi e^{-\beta \phi}$ can be added to the DMP's acceleration, where $\gamma, \beta > 0$, $\vect{R}$ is a rotation around the axis $\vect{k} = (\vect{p}_o - \vect{y}) \times \dvect{y}$ by angle $\pi/2$ and $\phi = \cos^{-1}( \frac{(\vect{p}_o - \vect{y})^T\dvect{y}}{|(\vect{p}_o - \vect{y})| |\dvect{y}|})$. 

\subsection{Additional Properties} \label{subsec:additional_properties}

Apart from reversibility, the proposed formulation decouples the DMP's effective stiffness and damping from the temporal scaling parameter $\tau$ (see \eqref{eq:GMP_imp_form} compared to \eqref{eq:DMP_imp} for the original one), which can affect the system's response in the presence of perturbations or other coupling terms. Moreover, training in the novel DMP requires only position measurement, whereas the original DMP  requires additionally velocity and acceleration measurements, which are usually noisy affecting the accuracy of the learning process. Hence,  computational load and memory resource demand is reduced with the proposed formulation. Apart from those advantages, two additional properties detailed below are supported. 

\textbf{\textit{Decoupled teaching of path and velocity profile:}}

Demonstrating a desired trajectory using kinesthetic guidance can prove to be quite cumbersome, as one has to pay attention to guide the robot accurately along the desired path, while at the same time imposing the desired speed of execution (velocity profile). This places a lot of cognitive load to the user and can deteriorate the quality of the demonstration. Moreover, there are tasks where accuracy in the demonstrated path is of great essence.  
A two phase learning approach was initially presented in \cite{Nemec_FrenetSeret_2017}, which uses the original DMP and learns the velocity profile and stiffness in the second phase.
With the proposed DMP, velocity profile  teaching can also be utilized in a second phase and is  easily realized, with a slight modification of the canonical system.
In particular, after the careful demonstration of the path (phase 1), in phase 2 the robot is under position control, following the output produced by  \eqref{eq:GMP_imp_form} which generates the path demonstrated during phase 1, but with different speed according to the following modified canonical system:
\begin{equation} \label{eq:GMP_vel_teach_x_dot}
\begin{cases}
    \ddot{x} &= -d_x \dot{x} + f_v \ , \ \text{for} \ x < 1  \\
    \ddot{x} &= \dot{x} = 0 \ , \ \text{for} \  x \ge 1
\end{cases}
\end{equation}
with $x(0)=1$, $\dot{x}(0)=0$, where $d_x>0$, $f_v = \vect{n}^T \vect{f}_{ext}$ with $\vect{n} = \frac{\partial \vect{y} / \partial x}{|\partial \vect{y} / \partial x|}$ being the unit vector pointing along the direction of the motion, $\vect{y}$ denotes the Cartesian position and $\vect{f}_{ext}$ is the external force provided by the robot's F/T sensor. 
At the end of this process, the recorded robot's Cartesian pose is used to retrain the DMP. 

\textbf{\textit{Bidirectional drivability along the path:}}

A similar modification of the canonical system can also be used for phase stopping and bidirectional drivability along the path:
\begin{equation} \label{eq:phase_stop_with_compliance}
    \ddot{x} = -d_x ( \dot{x} - \dot{x}_d) + f_v
\end{equation}
where $\dot{x}_d = s \frac{1}{\tau(1 + a_d ||\vect{F}_{ext}||)}$
with $x(0)=0$, $s=1$ for forward and $s=-1$ for reverse execution. Essentially, \eqref{eq:phase_stop_with_compliance} is obtained by combining \eqref{eq:phase_stop} and \eqref{eq:GMP_vel_teach_x_dot}. In the absence of external disturbances, it follows that $\dot{x} \rightarrow s/\tau$ and the motion evolves autonomously. When external forces are applied, phase stopping ensures that $\dot{x}_d \rightarrow 0$ and the motion evolves according the force along the path $f_v$.
Bidirectional drivability along the path can prove useful in many scenarios, e.g. to allow manual inspection of the executed trajectory, driving the robot back/forth along the learned path.



\section{Experimental Results} \label{sec:Experiments}

\begin{figure} [ht!]
    \centering
	\includegraphics[scale = 0.28]{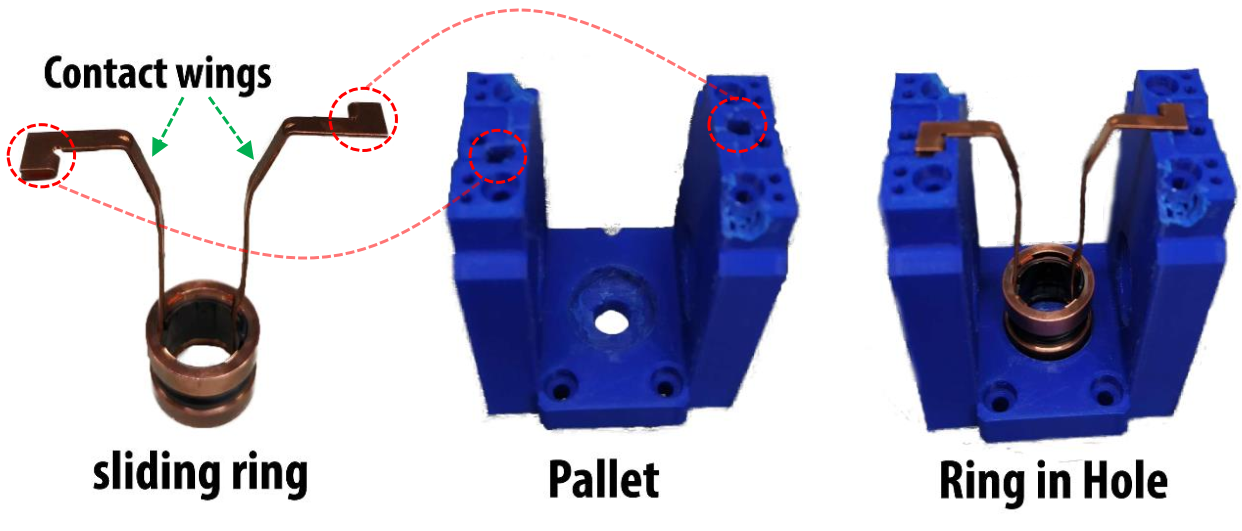}
	\caption{Ring in hole.}
	\label{fig:ring_in_hole}
\end{figure}

In this section we demonstrate the usefulness of the proposed DMP in a ring-in-hole (RiH) assembly. This assembly is inspired by a real case of a car starter assembly. The task involves the insertion of sliding rings with flexible contact wings shown in Fig. \ref{fig:ring_in_hole} into pallets consisting of a series of holes. 
We use the actual sliding rings and 3D printed pallets, from the CAD model of the actual pallets (Fig. \ref{fig:ring_in_hole}). The high flexibility and elasticity of the contact wings adds to the difficulty of the task as grasping the ring from the table and inserting it with the correct orientation is important because  the contact 'wings' of the ring should be properly aligned with the side walls of the pallet for the subsequent moulding operation to be successful.   

\vspace{-2em}
\begin{figure} [ht!]
    \centering
	\includegraphics[scale = 0.18]{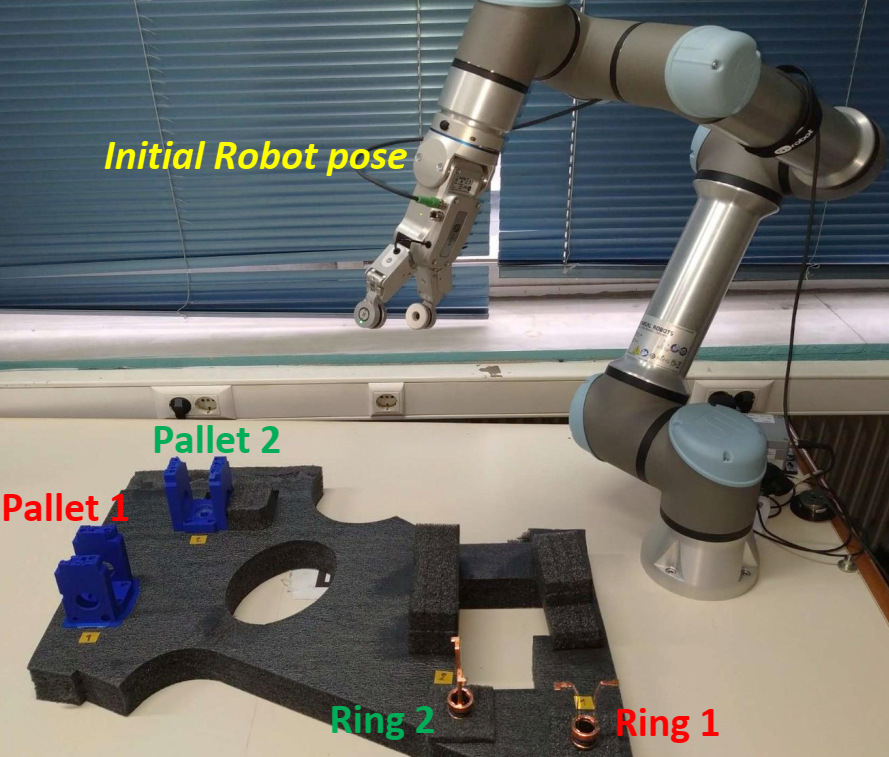}
	\caption{Ring in hole scene.}
	\label{fig:rih_scenario}
\end{figure} 

Based on this real case scenario, we have devised the task shown in Fig. \ref{fig:rih_scenario} for the insertion of 2 rings into corresponding pallets. We use a UR5e robot, which provides measurements of external wrenches, and an  RG2FT gripper attached at its wrist. 

During teaching, the geometric constraints due to the pallet's shape and the tight boundaries of the hole make it difficult for a human to demonstrate the insertion as he/she may have to wiggle the object before it fits properly in the hole and the sockets. This however deteriorates the quality of the demonstration. We resolve this problem by demonstrating the disassembly trajectory (shown in the video), starting with the ring already properly aligned and inserted into a pallet.
By flipping the collected data the encoding of the forward trajectory is then performed. The second phase of the training procedure (section \ref{subsec:additional_properties}) is then utilized to learn the velocity profile (shown in the video). Notice that in order for the DMP to generalize the learned motion pattern properly to new targets (pallet poses), the DMP is encoded w.r.t. the target pose (pallet pose).
Further notice that by demonstrating the reverse, the retraction motion will respect the geometric constraints.


The trained DMP is used for executing all the required motions of the task in sequence, starting from the initial robot pose, shown in Fig. \ref{fig:rih_scenario}, and moving to the pick/ insertion target poses.
Notice that all poses in  Fig. \ref{fig:rih_scenario}
are different from those employed during the demonstration. 
When the target is reached, the gripper closes (opens) for pick (insertion) and the robot retracts to its initial pose along the same (reverse) path. The latter is particularly important in the cluttered environment of the real case. To control the phase variable we use \eqref{eq:phase_stop_with_compliance}, applying also a dead-zone of $10N$ to $\vect{F}_{ext}$ when the distance to the target is below $1.5cm$ so that the robot can apply the required inserting force without stopping prematurely by the 'phase-stopping' mechanism. 

We showcase the usefulness of the reversibility and bi-directional drivability \eqref{eq:phase_stop_with_compliance} of the proposed DMP formulation in the context of manual trajectory inspection and automatic error recovery from faulty sensor measurements, exploring two specific scenarios. Specifically, during the execution of the first insertion, the user intervenes by driving the robot back/forth along the path to inspect the learned trajectory near the target. During the second insertion, a purposely wrong target is provided, which can be hypothetically attributed to perception errors of a vision sensor. This results in a collision with a surface generating a contact force and a premature stop, due to \eqref{eq:phase_stop_with_compliance}. Then we signal a partial retraction, 
utilizing \eqref{eq:phase_stop_with_compliance} for reverse execution until the phase variable $x$ reaches $0.2$, and then provide the correct target pose and resume the insertion, restoring $s$ to its initial value.
Notice that the detection for triggering this procedure could be automated by a higher level perception system, leveraging vision and/or force and proximity sensors. However, such detection schemes are beyond the scope of this work.

\begin{figure} [ht!]
    \centering
	\includegraphics[scale = 0.45]{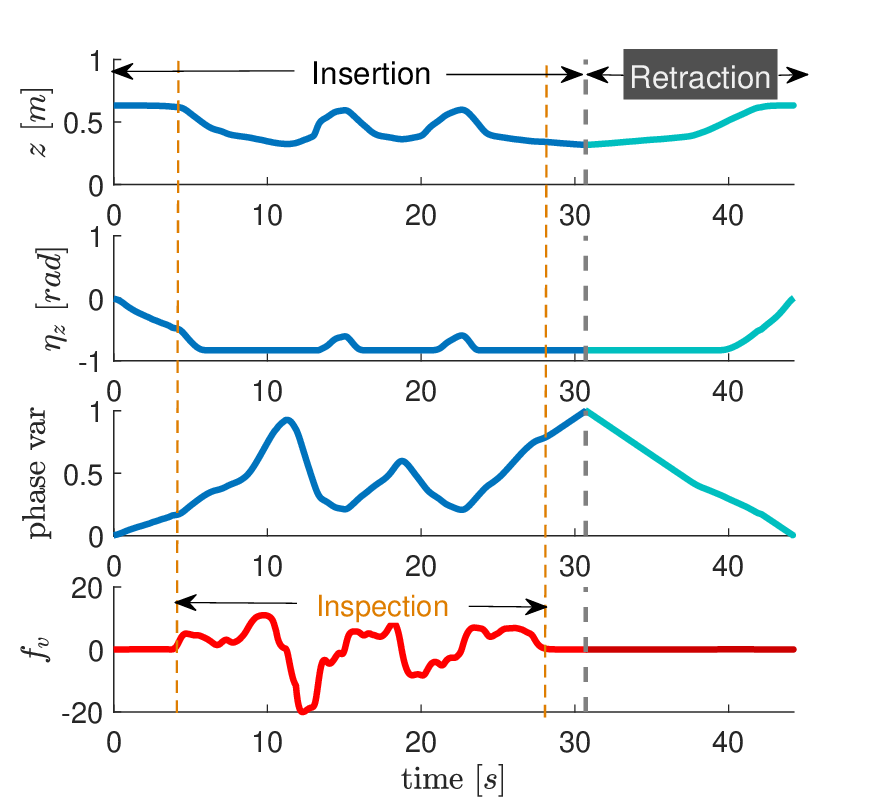}
	\caption{Ring 1 insertion and retraction: Position and orientation trajectories along $z$-axis with user inspection.}
	\label{fig:pick2_1D}
\end{figure} 

\begin{figure} [ht!]
    \centering
	\includegraphics[scale = 0.33]{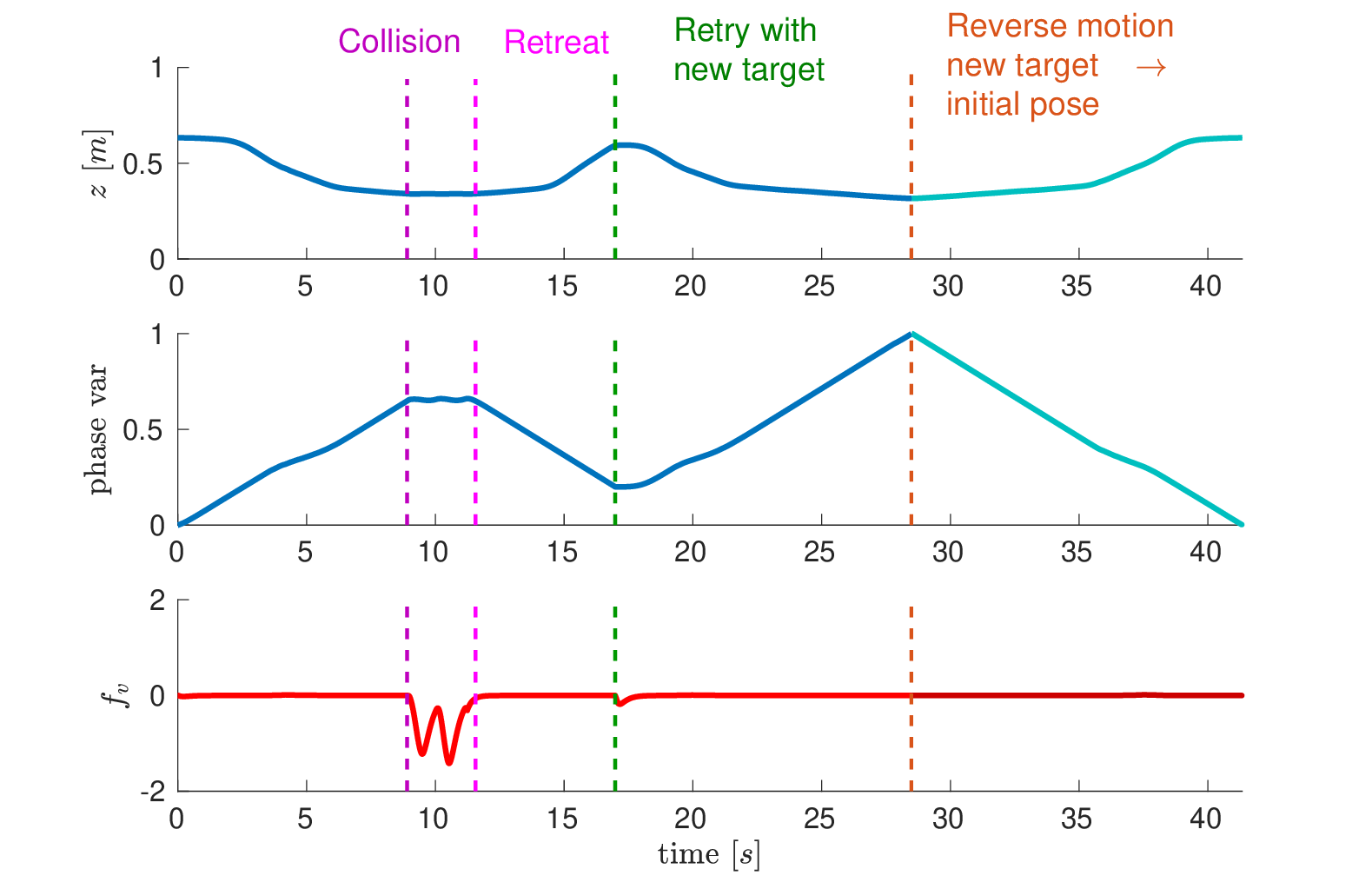}
	\caption{Ring 2 insertion: The initial target pose is wrong, leading to a collision. The robot retracts and then reattempts the insertion with the correct target.}
	\label{fig:place2_1D}
\end{figure} 

\begin{figure} [ht!]
    \centering
	\includegraphics[scale = 0.35]{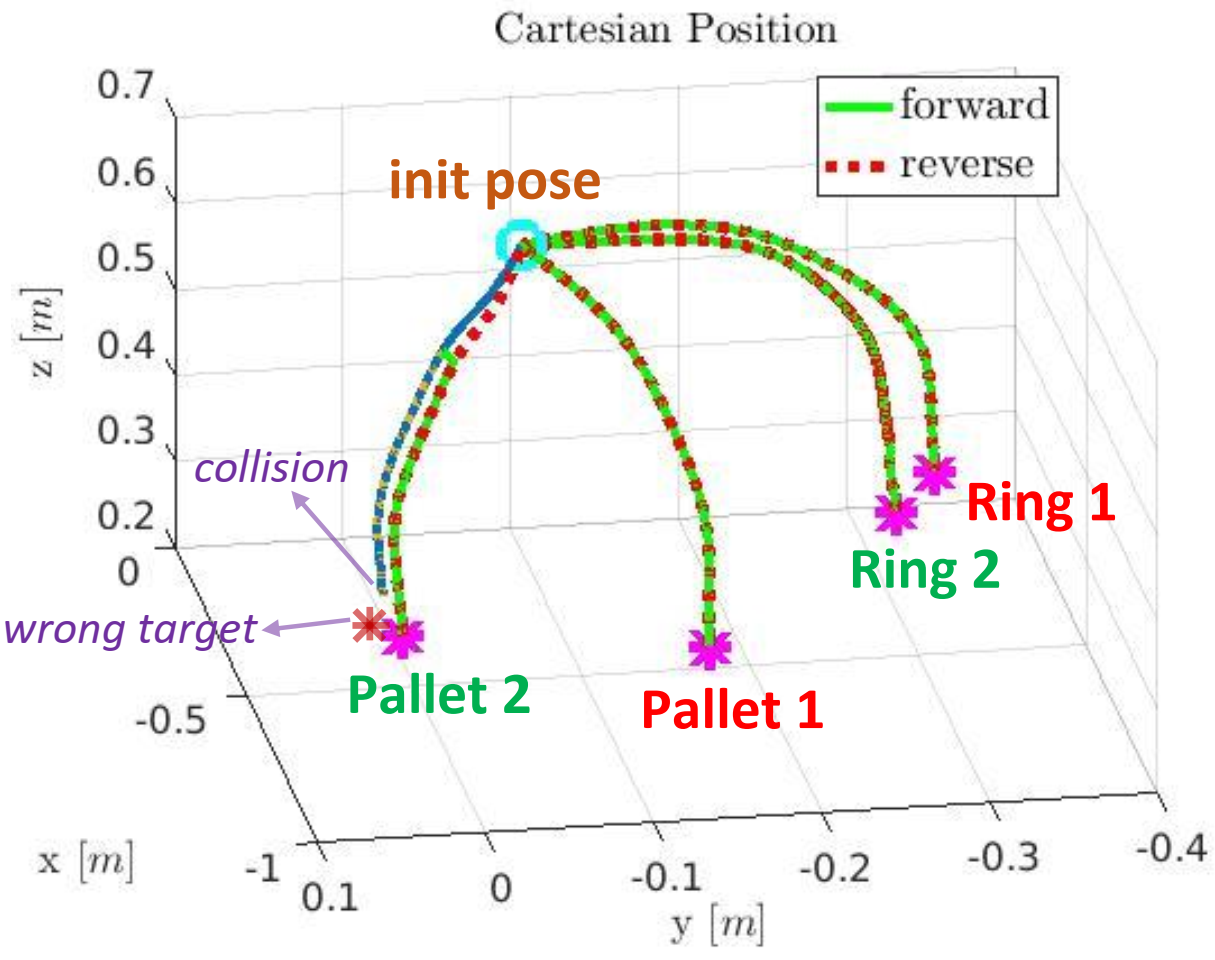}
	\caption{Position paths for all pick/insertions for forward motion with solid green line and reverse motion with red dotted line. During insertion at pallet 2, depicted with solid blue line, collision occurs due to having a wrong target. The robot retracts (dotted yellow line) and resumes insertion provided with the correct target. }
	\label{fig:exp_pos_3D_path}
\end{figure} 

\begin{figure} [ht!]
    \centering
	\includegraphics[scale = 0.4]{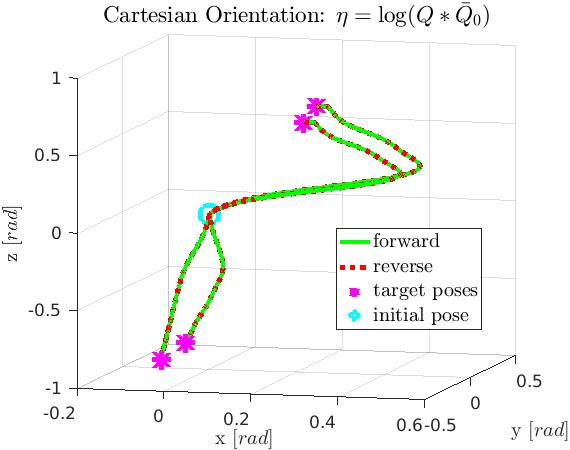}
	\caption{Orientation paths for forward and reverse motion for all pick/insertions, using $\eta = \log(\vect{Q}*\bvect{Q}_0)$}
	\label{fig:exp_orient_3D_path}
\end{figure}

Results are presented in Fig. \ref{fig:pick2_1D}-\ref{fig:exp_orient_3D_path}. During the insertion of $1$st ring the user intervenes (Fig. \ref{fig:pick2_1D}) to inspect the trajectory driving the robot back/forth along the path at $t=3.8 - 28$ sec. During the $2$nd ring insertion in Fig. \ref{fig:place2_1D}, a collision occurs at $t=8.9$ and the robot stops as it can be seen from the phase variable $x$ and the response along the $z$-direction followed by a retraction until $x = 0.2$ at $t=11.6$, and finally resumes the insertion to the correct target at $t=17$ sec. 
The $3$D Cartesian position and orientation paths for pick and insertion of both rings are plotted in Fig. \ref{fig:exp_pos_3D_path} and \ref{fig:exp_orient_3D_path}. The orientation is visualized using the quaternion logarithm $\vect{\eta} = \log(\vect{Q} * \bvect{Q}_0)$. Both the forward and reverse motion are constrained on the same path during the course of the entire experiment.

\section{Conclusions} \label{sec:Conclusions}

In this work, a reversible Dynamic Movement Primitive (DMP) formulation was proposed which ensures global asymptotic stability of the target in both forward and backward motion. All favourable properties of the  original DMP formulation are retained with a few additional desirable properties supported. The usefulness of the proposed formulation was showcased in RiH experiments.


\section*{Appendix A - Non-reversibility of the Original DMP and correspondence with the proposed DMP}

First we are going to examine the correspondence between the original and the proposed DMP. In particular, we will show that \eqref{eq:DMP_y_ddot} can written in the form of \eqref{eq:GMP_imp_form}.
Let $y_d(t_1), \partDer{y_d}{t_1}, \partDerSq{y_d}{t_1}$,  $t_1 \in [t_{0,d} \ t_{f,d}]$ be the trajectory  with duration $\tau_d = t_{f,d}-t_{0,d}$, goal $g_d$ and initial position $y_{d,0}$, that is generated by the trained forcing term \eqref{eq:DMP_f_s}. The forcing term of \eqref{eq:DMP_y_ddot} is thus given by:
\begin{equation} \label{eq:fs_learned}
    f_{s}(x) = \frac{ \tau^2_d \partDerSq{y_d}{t_1} - \alpha_z \beta_z (g_d - y_d) + \alpha_z \tau_d \partDer{y_d}{t_1}}{g_f(x)(g_d-y_{d,0})}
\end{equation}
\begin{equation} \label{eq:DMP_xd_dot}
    \tau_d \dot{x} = h(x)
\end{equation}
\begin{equation} \label{eq:DMP_xd_Ft}
    x = x_{0,d} + H(t_1 / \tau_d)  \ , \ t_1 \in [t_{0,d} \ t_{f,d}]
\end{equation}
where $H(t/\tau) \triangleq \int_{t_0/\tau}^{t/\tau} h(x(\sigma))d\sigma$.
The dependency of $y_d, \partDer{y_d}{t_1}, \partDerSq{y_d}{t_1}$ on $t_1$ is omitted for simplicity. 
Substituting \eqref{eq:fs_learned} in \eqref{eq:DMP_y_ddot} and after simple mathematical manipulations, we get:
\begin{equation} \label{eq:DMP_imp}
    \ddot{y} = \ddot{y}_{ref} - \frac{\alpha_z}{\tau}(\dot{y}-\dot{y}_{ref}) - \frac{\alpha_z \beta_z}{\tau^2}(y - y_{ref})
\end{equation}
\begin{align} 
    &y_{ref}(t) \triangleq k'_s(y_d(t_1) - y_{d,0}) + y_0 \label{eq:y_ref}\\
    &\dot{y}_{ref}(t) = k'_s k_t\partDer{y_d(t_1)}{t_1} \label{eq:y_ref_dot}\\
    &\ddot{y}_{ref}(t) = k'_s k_t^2 \partDerSq{y_d(t_1)}{t_1} \label{eq:y_ref_ddot}
\end{align}
where $k'_s = (g-y_0)/(g_d-y_{d,0})$ is the spatial and $k_t = \tau_d/\tau$ the temporal scaling factor.
Notice that  \eqref{eq:DMP_imp}-\eqref{eq:y_ref_ddot} is in the form of the proposed DMP \eqref{eq:GMP_imp_form}-\eqref{eq:GMP_y_x_ddot}, since $f_p(x)$ is trained based on $y_d(t_1)$ hence $f_p(x) \approx y_d(t_1)$ and in turn  $k_s \approx k'_s$, while $\dot{f}_p(x) \approx \dot{y}_d(t_1) = \partDer{y_d(t_1)}{t_1} \dot{t}_1 =  k_t \partDer{y_d}{t_1}$, which follows from $\dot{t}_1 = k_t$.
In particular, solving \eqref{eq:DMP_xd_Ft} for $t_1$ yields $t_1 = \tau_d H^{-1}(u)$, with $u = x - x_{0,d}$. Its time derivative is $\dot{t}_1 = \tau_d\partDer{H^{-1}(u)}{u} \dot{u} = \tau_d \left( h(x) \right)^{-1} \dot{x}$ by applying the inverse function Theorem (see theorem 2.9 in \cite{Nonlinear_Control_Systems_Marquez}).
Substituting $\dot{x}$ from \eqref{eq:DMP_x_dot} we get $\dot{t}_1 = k_t$.

\begin{remark}
To validate that \eqref{eq:y_ref_dot}, \eqref{eq:y_ref_ddot} are the first and second order time derivatives of $y_{ref}$ notice that:
\begin{equation} \label{eq:y_ref_dot_2}
    \dot{y}_{ref}(t) = k'_s \frac{d {y}_d(t_1)}{dt} = k'_s \partDer{y_d}{t_1} \dot{t}_1 = k'_s k_t \partDer{y_d}{t_1}
\end{equation}
where we used the fact that $\dot{t}_1 = k_t$. Accordingly, taking the time derivative of $\dot{y}_{ref}(t)$ it follows easily that $\ddot{y}_{ref}(t) = k'_s k_t \frac{d}{dt}\left( \partDer{y_d}{t_1} \right) = k'_s k_t \partDerSq{y_d}{t_1} \dot{t}_1 = k'_s k_t^2 \partDerSq{y_d}{t_1}$.
\end{remark}

Despite the similarity in the form between the proposed and the original DMP, their main difference is in the encoding term which in the proposed formulation utilizes only the demonstrated position trajectory while in the original, a combination of the demonstrated position velocity and acceleration trajectories are utilized (see \eqref{eq:fs_learned}). The consequence is the non-reversibility of the original formulation as we prove in the following.
Specifically, using the reverse canonical system $\tau \dot{x} = -h(x)$ of \eqref{eq:DMP_x_dot} and following the procedure as for deriving \eqref{eq:DMP_imp}, we can conclude that \eqref{eq:DMP_y_ddot} reduces to:
\begin{equation} \label{eq:DMP_imp_rev}
    ( \ddot{y} - \ddot{y}_{rev} ) + \frac{\alpha_z}{\tau}(\dot{y} - \dot{y}_{rev}) + \frac{\alpha_z \beta_z}{\tau^2}(y - y_{rev}) = -2\frac{a_z}{\tau}\dot{y}_{rev}
\end{equation}
where $y_{rev} = k_s(y_d(\tau_d H^{-1}(x-x_{0,d})) - y_{d,0}) + y_0$
hence the DMP output $y, \ \dot y, \ \ddot y$ will not track faithfully the reverse trajectory due to the existence of the term $-2\frac{a_z}{\tau}\dot{y}_{rev}(t)$ at the right hand-side of \eqref{eq:DMP_imp_rev}, which acts as a disturbance. It can be easily verified by simulations that the reverse DMP exhibits tracking errors and fails to reach the target at the designated time duration.

\section*{Appendix B - Unit Quaternion Preliminaries}

Given a rotation matrix  $\vect{R}\in SO(3)$, an orientation can be expressed in terms of the unit quaternion $\vect{Q} \in \mathbb{S}^{3}$ as $\vect{Q}=[w \ \vect{v}^T]^T = [\cos(\theta) \ \sin(\theta) \vect{k}^T]$,
where $\vect{k} \in \mathbb{R}^3$, $2\theta \in [0 \ 2\pi)$ are the equivalent unit axis - angle representation.
The quaternion product between the unit quaternions $\vect{Q}_1$, $\vect{Q}_2$ is denoted as $\vect{Q}_1*\vect{Q}_2$.
The inverse of a unit quaternion is equal to its conjugate which is
$\vect{Q}^{-1} = \bar{\vect{Q}} = [w \ -\vect{v}^T]^T$.
The logarithmic $\vect{\eta} = \log(\vect{Q})$ and exponential $\vect{Q} = \exp(\vect{\eta})$ mappings $\log: \ \mathbb{S}^3 \rightarrow \mathbb{R}^3$, $\exp: \ \mathbb{R}^3 \rightarrow \mathbb{S}^3$ respect the manifold's geometry and are defined as follows:
\begin{equation} \label{eq:quatLog}
    \log(\vect{Q}) \triangleq 
        \left\{
            \begin{matrix}
                2\cos^{-1}(w)\frac{\vect{v}}{||\vect{v}||}, \ |w|\ne1 \\
                [0,0,0]^T, \ \text{otherwise}
            \end{matrix}
        \right.
\end{equation}
\begin{equation} \label{eq:quatExp}
    \exp(\vect{\eta}) \triangleq 
        \left\{
            \begin{matrix}
                [\cos(||\vect{\eta}/2||), \sin(||\vect{\eta}/2||)\frac{\vect{\eta}^T}{||\vect{\eta}||}]^T, \ ||\vect{\eta}|| \ne 0 \\
                [1,0,0,0]^T, \ \text{otherwise}
            \end{matrix}
        \right.
\end{equation}
Denoting by $\vect{\Omega} = [0 \ \vect{\omega}^T]^T$, the time derivatives of $\vect{\eta}$,   are related to $\vect{\omega}$, $\dvect{\omega}$ as follows \cite{DMP_orient_Koutras}:
\begin{equation} \label{eq:dq_omega}
    \vect{\Omega} = 2 (\vect{J}_{\eta}\dvect{\eta})*\bvect{Q}
\end{equation}
\begin{equation} \label{eq:dvRot_ddq}
    \dvect{\Omega} = 2 (\vect{J}_{\eta}\dvect{\eta} + \vect{J}_{\eta}\ddvect{\eta})*\bvect{Q} - \frac{1}{2}\begin{bmatrix} ||\vect{\omega}||^2 \  \vect{0}_{1 \times 3}  \end{bmatrix}^T
\end{equation}
\begin{equation*} \label{eq:J_Q}
    \vect{J}_{\eta} = \frac{1}{2}
    \begin{bmatrix}
        -\sin(\theta) \vect{k}^T \\
        \frac{\sin(\theta)}{\theta}(\vect{I}_3-\vect{k}\vect{k}^T) + \cos(\theta)\vect{k}\vect{k}^T
    \end{bmatrix}
\end{equation*}

\newpage

\bibliographystyle{IEEEtran}
\bibliography{mybib}

\addtolength{\textheight}{-12cm}   



\end{document}